RESEARCH ARTICLE

# Title: BDPM：A Machine Learning-Based Feature Extractor for Parkinson's Disease Classification via Gut Microbiota Analysis

Bo Yu*, Zhixiu Hua, Bo Zhao

**Abstract:** *Background:* Parkinson's disease remains a major neurodegenerative disorder with high misdiagnosis rates, primarily due to reliance on clinical rating scales. Recent studies have demonstrated a strong association between gut microbiota and Parkinson's disease, suggesting that microbial composition may serve as a promising biomarker. Although deep learning models based on gut microbiota show potential for early prediction, most approaches rely on single classifiers and often overlook inter-strain correlations or temporal dynamics. Therefore, there is an urgent need for more robust feature extraction methods tailored to microbiome data.

*Methods:* We proposed BDPM (A Machine Learning-Based Feature Extractor for Parkinson's Disease Classification via Gut Microbiota Analysis). First, we collected gut microbiota profiles from 39 Parkinson's patients and their healthy spouses to identify differentially abundant taxa. Second, we developed an innovative feature selection framework named RFRE (Random Forest combined with Recursive Feature Elimination), integrating ecological knowledge to enhance biological interpretability. Finally, we designed a hybrid classification model to capture temporal and spatial patterns in microbiome data.

*Results:* BDPM achieved excellent performance in distinguishing Parkinson's patients from controls, with mean accuracy, precision, recall, F1 score, AUC, and ROC of 0.97, 0.97, 0.95, 0.96, and 0.97, respectively. The model effectively leveraged differences in gut microbiota composition between groups, providing insights into the Brain-Gut-Microbiome Axis.

*Conclusion:* BDPM introduces a novel feature extraction pipeline specifically designed for microbiome data, offering improved accuracy and interpretability for Parkinson's disease classification. This work highlights the potential of integrative machine learning approaches in advancing early diagnosis and prevention strategies for neurodegenerative disorders.

## 1. INTRODUCTION

Parkinson's disease (PD), also known as "tremor paralysis," is a common neurodegenerative disorder primarily affecting the elderly population[1]. Projections indicate that by 2023, China will account for approximately half of all global PD cases[2], highlighting its growing public health significance. Notably, PD is no longer confined to older adults; increasing evidence suggests a trend toward earlier onset[3]. The pathogenesis of PD is closely associated with the degeneration of dopaminergic neurons in the substantia nigra[4]. Currently, diagnosis relies heavily on clinical rating scales, which are not only time-consuming but also subjective, as they depend on the physician's experience and patient self-reporting[5]. This lack of objective biomarkers contributes to high rates of early misdiagnosis[6]. PD progression can be divided into three stages: preclinical, prodromal, and clinical[7].

*Address correspondence to this author at the Department of Software Engineering, Faculty of Computer Science and Technology, Harbin University of Science and Technology, E-mail: yubo@hrbust.edu.cn, Harbin, China

Research indicates that the prodromal phase may last up to 20 years before clinical symptoms emerge[8]. Although PD remains incurable[9], early detection—particularly during the prodromal stage—can significantly delay disease progression. Thus, developing objective and efficient diagnostic tools is of critical importance. In recent years, mounting evidence has highlighted the role of gut microbiota in PD. Studies suggest a potential "Brain-Gut-Microbiome Axis," where in pathological changes may originate in the enteric nervous system before spreading to the brain[10]. For instance, Sangjune Kim et al. demonstrated that α-synuclein pathology can propagate via the "Brain-Gut Axis" using the mouse model[11]. Non-motor symptoms such as constipation often appear years before motor symptoms, further supporting the idea that gut microbiota may serve as an early predictive biomarker for PD[12]. Exploring the link between gut microbiota and PD offers new perspectives for early diagnosis and may reveal novel therapeutic targets. These findings could pave the way for more effective treatment strategies. Machine learning (ML) has shown great promise in PD detection, particularly in speech analysis[13], gait assessment, and medical imaging[14-17]. However, most existing studies focus on conventional biomarkers such as blood, urine, or protein samples[18-20]. Gut microbiota-based models remain underexplored despite their non-invasive nature and biological relevance. Given the high-dimensional and sparse characteristics of microbiome data, there is a pressing need for feature extraction methods tailored to this type of data. To address this gap, we proposed BDPM. BDPM integrates ecological knowledge into feature selection through an innovative RFRE framework and employs a temporal-spatial dual cascade classification network to enhance prediction accuracy. Using ten-fold cross-validation, BDPM achieved a mean accuracy (ACC), precision, recall, F1 score, and AUC of 0.97, 0.97, 0.95, 0.96, and 0.97, respectively—outperforming existing methods. This work provides a novel, non-invasive approach for PD prediction based on gut microbiota, offering the potential for early diagnosis and improved patient outcomes.

## 2. MATERIALS AND METHODS

### 2.1. Data collection and dataset construction

The data used in this study were obtained from a cross-sectional study on the gut microbiota of Parkinson's disease (PD) patients in Central China[21], including 39 matched pairs of PD patients and healthy controls. The diagnosis was based on the Movement Disorder Society (MDS) clinical diagnostic criteria for PD (2015)[22]. According to these criteria, a definitive diagnosis requires the presence of parkinsonism (e.g., bradykinesia) combined with either resting tremor or rigidity. For sample collection, fecal specimens were immediately frozen at -80 °C after collection to preserve microbial DNA stability. Total DNA was subsequently extracted using the MetaHIT standardized protocol and quantified using a Qubit fluorometer. Following library construction, metagenomic sequencing was performed. Raw sequencing data were first evaluated for quality using FastQC, followed by trimming of low-quality reads and removal of host-derived contaminants. Taxonomic profiling and estimation of relative species abundance were conducted using MetaPhlAn 2.0.

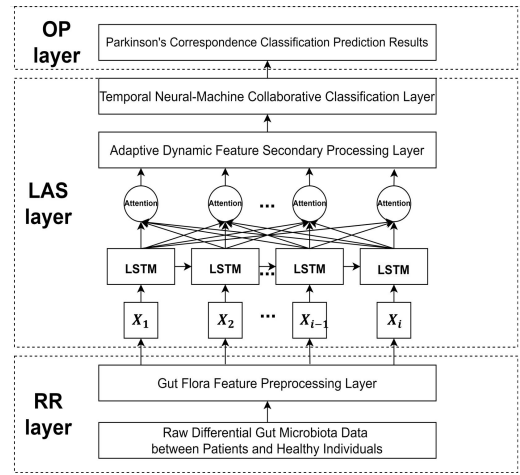

**Fig. (1).** The overall algorithm framework of BDPM. BDPM is composed of three layers: first, feature preprocessing is performed at the RR layer; subsequently, the proposed algorithm is applied for training and classification at the LAS layer; finally, the prediction results are obtained at the OP layer.

The final species abundance values used for downstream analysis were calculated by multiplying the relative abundance by the total read count per sample and rounding to the nearest integer.

### 2.2. Parkinson's prediction model

#### 2.2.1. Overall network framework

The gut microbiota has been shown to influence PD via the brain-gut axis mechanism. Based on this relationship, we proposed a novel method for PD prediction by leveraging differences in gut microbial composition between patients and healthy individuals. Given the high dimensionality and complex interactions within the gut microbiome, as well as the limited sample size typical of such studies, we introduced a three-stage hierarchical framework that integrates ecological knowledge with machine learning to address the "high-dimensional, small-sample" challenge.

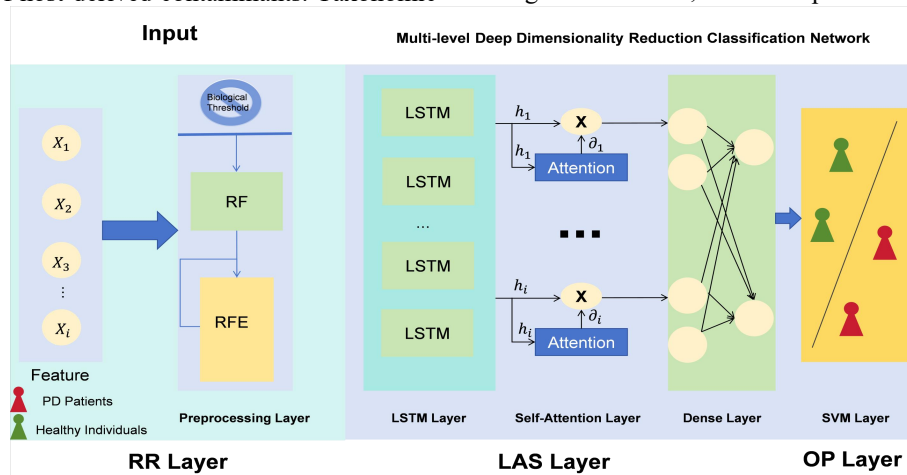

**Fig. (2).** This is an overall overview of the process for predicting Parkinson's disease using gut microbiota data with BDPM. The diagram provides an overview of the principles and workflow of BDPM. BDPM integrates machine learning and biological knowledge in the preprocessing and training stages of gut microbiota data. The prediction is mainly based on the differences in gut microbiota between patients and their healthy spouses, offering an innovative perspective for Parkinson's disease research.

The proposed BDPM framework is illustrated in Fig.1 and consists of four modules: RR (RFRE Preprocessing), LA (LSTM-Attention), LS (LA-SVM), and OP (Output), organized into a three-tier architecture. The first layer (RR) performs biologically informed feature selection and dimensionality reduction to meet training requirements. In the second layer (LA), the LSTM-Attention module adaptively refines temporal features. Building upon LA, the LS layer further combines the extracted features with an SVM-based classification network. Finally, the OP layer generates the final prediction.

BDPM is trained using the Adam optimizer, combining the advantages of sequential modeling with the robustness of traditional machine learning. This design enhances the interpretability of microbial features while effectively mitigating the risk of overfitting in small-sample scenarios, thereby significantly improving model performance. The detailed algorithm workflow is presented in Fig.

### 2.2.2. Bio-deep dimensionality reduction preprocessing approach

During data preprocessing, many studies use single filters or ignore ecological principles. While simple and intuitive, these approaches often lack biological relevance, leading to reduced accuracy. To address this, we propose a bio-deep dimensionality reduction method within our BDPM framework.

We developed RFRE (Random Forest and Recursive Feature Elimination) to integrate ecological niche theory with machine learning. Unlike traditional methods, RFRE combines ecological insights with advanced feature selection for interdisciplinary analysis. First, we applied a minimum abundance threshold to remove low-abundance noise taxa and retain core taxa (>0.005% of the highest abundance). This formed the first level of biological filtering. In the second stage, guided by machine learning, both RF and RFE were used for feature compression.

The gut microbiome is highly diverse, with abundance differences spanning orders of magnitude. Species interactions further complicate their quantification. Random Forest excels at handling large datasets, identifying important features, and preventing overfitting. RFE iteratively removes less relevant features, forming dynamic importance subsets and adapting to non-linear relationships. By combining RFE and Random Forest, we created an 'evaluate-remove-evaluate' loop that enhances performance while maintaining low feature dimensionality.

The RFRE involves the following feature selection steps:

Step 1: Key microbial taxa associated with PD are selected based on the flora difference table, while irrelevant taxa are removed to reduce noise and improve model interpretability and generalization. To better assess feature importance, the original data matrix is transposed so that each column represents a sample.

Step 2: To reduce dimensionality and remove low-abundance noise, we compute the total abundance of each taxon across all samples. A threshold is defined as a percentage of the highest observed abundance ( = 0.005%). Only taxa exceeding this threshold are retained. The calculation is as follows:

$$TotalAbundance_j = \sum_{i=1}^{n} X_{ij} \quad (1)$$

$$ValidCols = \{j | TotalAbundance_j > \tau \cdot max(TotalAbundance)\}$$

Where $TotalAbundance_j$ denotes the total abundance of taxon $j$, $X_{ij}$ is the abundance of taxon $j$ in sample $i$, $n$ is the total number of samples, $\tau$ is a threshold ratio employed to determine whether a strain is screened or not, and $ValidCols$ records the indices of selected taxa.

Step 3: After biological filtering, RF is used to evaluate feature importance based on Gini impurity reduction:

$$\text{Gini}(D) = 1 - \sum_{k=1}^{n} p_k^2 \quad (2)$$

$$\text{Importance}(A) = \frac{1}{T}\sum_{t=1}^{T}\sum_{splits A} \Delta \text{Gini}_t \quad (3)$$

where $D$ is the current node dataset, $K$ is the number of classes (PD = 1, control = 0), is the proportion of class, $T$ is the number of trees, and $\Delta Gini_t$ is the Gini reduction by feature A in tree t.

Subsequently, RFE iteratively removes the least important features until 40 final features remain:

$$Ranking(A) = argsort(-Importance(A)) \quad (4)$$

$$F_{sorted} = argsort(-I_j) \quad (5)$$

$$F_{selected} = \{f_j | Importance(f_j) \in Top40(I_j)\} \quad (6)$$

Where $F_{sorted}$ contains the indices of the top 40 selected taxa, $I_j$ represents the global feature importance scores of $f_j$, $argsort()$ refers to the sorted importance scores, and $F_{selected}$ indicates the final 40 selected microbial taxa.

Step 4: To eliminate scale differences between high- and low-abundance taxa and prevent bias during training, the selected features are normalized:

$$X_{scaled} = \frac{X - min(X)}{max(X) - min(X)} \quad (7)$$

This improves convergence speed in LSTM and enhances stability in SVM, while also reducing the impact of outliers. In this context, $X$ is the original abundance value of the microbial taxa before normalization, and $X_{scaled}$ is the normalized feature value.

Step 5: Finally, disease labels (PD/healthy) are added to the processed features to ensure data integrity and avoid accidental label modification during preprocessing, facilitating subsequent feature-label separation.

### 2.2.3 Time-Space Downscaling and Classification

The preprocessed data has undergone bio-deep dimensionality reduction. To obtain the final classification results, we introduced the LAS layer — a temporal-spatial collaborative dimensionality reduction and classification module based on LSTM with Self-Attention for Temporal-Spatial Dimensionality Reduction and Classification via SVM (LAS).

In this architecture, the LSTM component captures temporal dependencies among microbial taxa, while the self-attention mechanism dynamically weights these relationships to enhance feature interaction. Given the limited number of training samples in Parkinson's disease detection, which makes single classifiers prone to overfitting, the BDPM framework incorporates a neural-spatial synergy mechanism and employs an LSTM-SVM cascade model as the final classifier.

In our LA module, the attention mechanism is applied after the LSTM layer to construct a fully connected layer. Specifically, LSTM first extracts temporal dependencies among species; then, the attention mechanism focuses on relevant interactions. The resulting features are compressed through a fully connected layer and subsequently fed into SVM for classification. This design effectively integrates temporal-spatial structure with the robustness of machine learning under small-sample conditions, thereby enhancing generalization while ensuring classification accuracy.

LSTM is a variant of RNN capable of capturing long-term dependencies, overcoming the limitations caused by short-term memory in traditional RNNs. Its core components include a triple-gate mechanism (forget gate, input gate, and output gate) and a cell state that preserves information across time steps. The forget gate determines which information from the previous cell state should be discarded, mapping the hidden state through a Sigmoid function to values between 0 and 1 — where 0 indicates complete forgetting then 1 indicates full retention. The input gate controls how much new information is added to the current cell state, using a combination of a Sigmoid gate and a Tanh-transformed candidate value. Finally, the output gate regulates what part of the updated cell state is passed to the next time step. Through this mechanism, LSTM effectively addresses the gradient vanishing problem in RNNs and enables effective temporal dimensionality reduction.

The formulae for temporal relation extraction are as follows:

1 Forget Gate：
$$f_t = \sigma(W_f \cdot [h_{t-1}, x_t] + b_f) \qquad (8)$$

2 Input Gate & Candidate Values：
$$\widetilde{C}_t = tanh(W_c \cdot [h_{t-1}, x_t] + b_c) \qquad (9)$$
$$i_t = \sigma(W_i \cdot [h_{t-1}, x_t] + b_i) \qquad (10)$$

3 Cell state update：
$$C_t = f_t \odot C_{t-1} + i_t \odot \widetilde{C}_t \qquad (11)$$

4 Output Gate & Hidden State：
$$o_t = \sigma(W_o \cdot [h_{t-1}, x_t] + b_o) \qquad (12)$$
$$h_t = o_t \odot tanh(C_t) \qquad (13)$$

The self-attention mechanism, also known as internal attention, is a method that associates different positions within a single sequence to compute its contextual representation. In deep learning, it is widely used to dynamically adjust feature weights through selective attention, thereby emphasizing important information while suppressing less relevant details.

This mechanism is typically applied in global modeling and can explicitly capture long-range dependencies between elements in a sequence. When integrated with LSTM, it further enhances the model's capacity to handle long-distance dependencies, offering greater flexibility.

In this study, we embed the self-attention mechanism between the LSTM layer and the fully connected layer to achieve depth-core dimensionality reduction, completing the three-level architecture. By processing 240-dimensional hidden states, the mechanism outputs weighted temporal features, improving both model accuracy and computational efficiency.

The attention mechanism is implemented using three components: *Query (Q)*, *Key (K)*, and *Value (V)*. The corresponding matrices and weight calculation formulas are as follows:

$$Q = XW^Q, \; K = XW^K, \; V = XW^V \qquad (14)$$
$$\text{Attention}(Q, K, V) = \text{softmax}\left(\frac{QK^\top}{\sqrt{d_k}}\right)V \qquad (15)$$

The fully connected layer maps high-dimensional temporal features, obtained after capturing temporal dependencies, into a lower-dimensional space. This step retains discriminative features suitable for SVM input while eliminating dimensional redundancy. The transformation is defined as follows:

$$Output = W \cdot x + b \qquad (16)$$

Where denotes the weight matrix, is the bias vector, and represents the input feature.

**Table 1.** RFRE Preprocessing Results. Table 1 enumerates the top 20 most significant microbial species in the dataset, along with their importance scores.

| Species Name | Score | Species Name | Score |
|---|---|---|---|
| Bifidobacterium_dentium | 0.0685 | Lactobacillus_salivarius | 0.0222 |
| Bilophila_unclassified | 0.0600 | Anaerotruncus_colihominis | 0.0214 |
| Ruminococcaceae_bacterium_D16 | 0.0509 | Erysipelotrichaceae_bacterium_2_2_44A | 0.0208 |
| Alistipes_putredinis | 0.0423 | Rothia_dentocariosa | 0.0200 |
| Alistipes_indistinctus | 0.0421 | Clostridiales_bacterium_1_7_47FAA | 0.0187 |
| Subdoligranulum_unclassified | 0.0383 | Lachnospiraceae_bacterium_9_1_43BFAA | 0.0186 |
| Scardovia_wiggsiae | 0.0376 | Leuconostoc_pseudomesenteroides | 0.0163 |
| Clostridium_leptum | 0.0360 | Butyricimonas_synergistica | 0.0157 |
| Clostridium_hathewayi | 0.0343 | Clostridium_sp_L2_50 | 0.0149 |
| Lachnospiraceae_bacterium_3_1_57FAA_CT1 | 0.0342 | Alistipes_sp_AP11 | 0.0128 |
| Peptostreptococcaceae_noname_unclassified | 0.0334 | candidate_division_TM7_single_cell_isolate_TM7b | 0.0222 |
| Clostridium_citroniae | 0.0325 | Erysipelotrichaceae_bacterium_21_3 | 0.0214 |
| Gemella_haemolysans | 0.0323 | Streptococcus_pasteurianus | 0.0208 |
| Bilophila_wadsworthia | 0.0318 | Bacteroides_sp_3_1_19 | 0.0200 |
| Subdoligranulum_variabile | 0.0274 | Subdoligranulum_sp_4_3_54A2FAA | 0.0187 |
| Clostridium_symbiosum | 0.0261 | Olsenella_unclassified | 0.0186 |
| Parabacteroides_goldsteinii | 0.0259 | Scardovia_unclassified | 0.0163 |
| Bacteroides_coprocola | 0.0251 | Blautia_hydrogenotrophica | 0.0157 |
| Oxalobacter_formigenes | 0.0236 | Fusobacterium_varium | 0.0149 |
| Clostridium_asparagiforme | 0.0234 | Oscillibacter_sp_KLE_1728 | 0.0128 |

Support Vector Machine (SVM) is a classical supervised learning model widely used for binary classification. Its core principle lies in finding an optimal hyperplane — also known as the decision boundary — in the high-dimensional feature space that maximizes the margin between two classes. SVM demonstrates strong robustness against overfitting, particularly in high-dimensional settings.

The optimization objective, constraint function, and decision function of SVM are formulated as follows:

$$\min_{w,\xi} \frac{1}{2} ||w||^2 + C \sum_i \xi_i \quad (17)$$

$$s.t. \, y_i(w \cdot x_i + b) \geq 1 - \xi_i, \, \xi_i \geq 0 \quad (18)$$

$$f(x) = sign\left(\sum_i \alpha_i y_i K(x_i, x) + b\right) \quad (19)$$

Where $w$ is the weight vector, $C$ is the regularization parameter, $\xi_i$ denotes the slack variables, $b$ is the bias term, $\alpha_i$ represents the Lagrange multipliers, and $K$ is the kernel function.

## 3. EXPERIMENTA

### 3.1. Training of models

Given the limited size of the dataset, we randomized the data to mitigate potential distributional bias. We conducted 10-fold cross-validation to evaluate model performance comprehensively. Specifically, we partitioned the dataset into 10 equal subsets. In each iteration, one subset served as the test set, while the remaining nine were combined to form the training set.

We carried out comparative experiments across multiple models using identical hyperparameter settings. We configured the training process for 500 epochs with an initial learning rate of 0.001. Additionally, we applied a batch size of 8 and used the Adam optimizer (Adaptive Moment Estimation) for parameter updates.

### 3.2. RFRE Preprocessing Results

During the biotic-depth dimensionality reduction preprocessing, we removed redundant information while preserving the relationships among microbial taxa. By introducing the concept of ecological thresholds and applying RFRE for feature selection, we retained key features to enhance model performance. Table 1 presents the top 40 selected microbial taxa along with their contribution scores.

As shown in Table 1, the top five core taxa contributed the most to model training. For clarity, we also list the importance scores of the top 20 taxa in the same table.

To reduce the black-box nature of the model and enhance its interpretability, we visualized the effects of the top 10 most influential microbial taxa on Parkinson's disease prediction, as well as selected interactions among these taxa, as shown in Fig. 3.

**Fig. (3).** The following partially visualizes the contribution rates of gut microbiome features and their interrelationships using SHAP analysis. This demonstrates the complex interactions among gut microbiomes, which necessitated multiple optimization attempts in BDPM before finalizing the methodology and parameters.

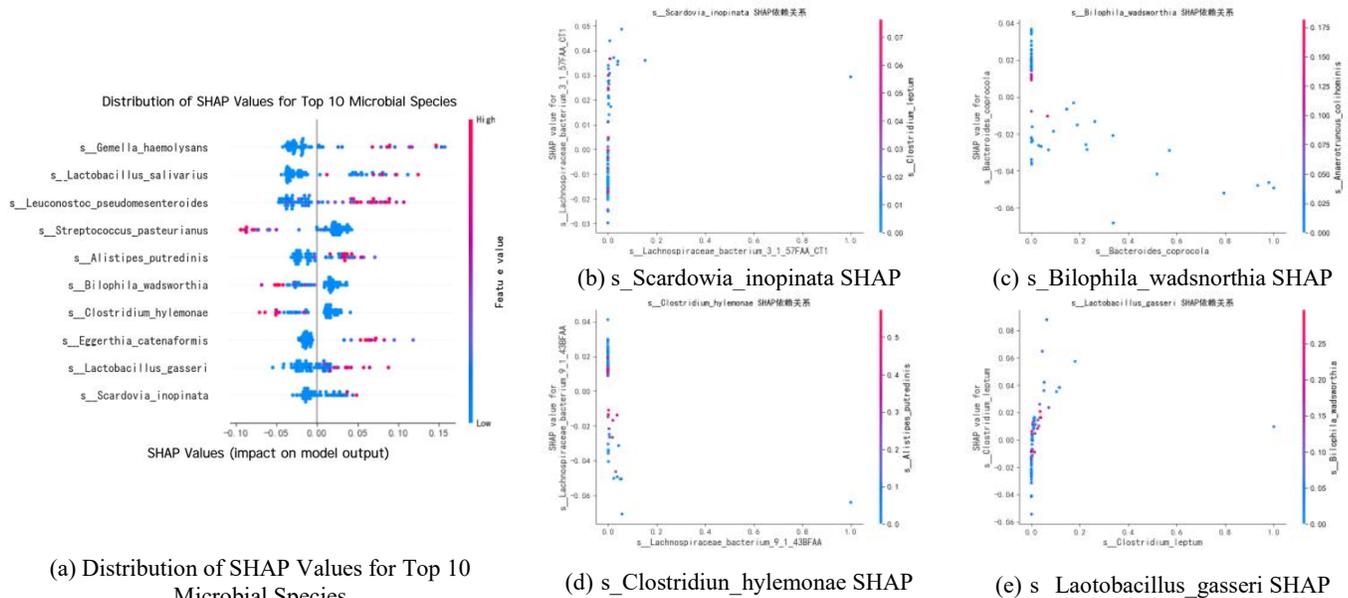

(a) Distribution of SHAP Values for Top 10 Microbial Species
(b) s_Scardowia_inopinata SHAP
(c) s_Bilophila_wadsnorthia SHAP
(d) s_Clostridiun_hylemonae SHAP
(e) s_Laotobacillus_gasseri SHAP

The visualization is based on SHAP (SHapley Additive exPlanations) values. In the plot, each dot represents a sample. The horizontal axis shows the SHAP value — indicating the impact of a specific taxon on the model output — while the vertical axis lists different taxa. The color reflects the abundance of the corresponding taxon in the sample: blue indicates low abundance, and red indicates high abundance.

For instance, Gemella haemolysans exhibits negative SHAP values across most samples, suggesting a protective effect in general. However, in a few samples with high abundance (red dots), the SHAP values are positive (0.05–0.15), indicating that this bacterium may have an opposite effect under certain high-abundance conditions.

A wider distribution range of SHAP values implies a stronger influence of the corresponding feature on model predictions, allowing us to intuitively understand how each taxon affects the outcome. Gemella haemolysans and Lactobacillus salivarius appear to be protective at normal abundance levels, which aligns with previous findings showing the beneficial effects of lactic acid bacteria on the nervous system. In contrast, Streptococcus pasteurianus, Bilophila Wadsworth, and Clostridium hylemonae may be associated with an increased risk of Parkinson's disease. These taxa could potentially affect the nervous system through the production of specific metabolites or by modulating intestinal inflammation.

Moreover, several taxa exhibit abundance-dependent effects, highlighting the importance of microbial balance over individual species. Notably, Lactobacillus gasseri and others show substantial inter-individual variation, suggesting that the gut microbial mechanisms underlying Parkinson's disease may differ across individuals.

These findings offer valuable insights into the relationship between gut microbiota and PD. They also hold promise for the development of microbiome-based diagnostic biomarkers and personalized microbial intervention strategies.

### 3.3. Performance Evaluation

Given the limitations of single evaluation metrics in representing and interpreting model performance, this study establishes a five-dimensional evaluation framework to provide a more comprehensive assessment.

We selected Accuracy (Acc) as the baseline metric, which reflects overall predictive performance by calculating the proportion of correctly classified samples in the test set. To evaluate classification reliability, Precision and Recall were introduced. Precision measures the proportion of true positive predictions among all positive predictions, indicating the reliability of identifying healthy cases. Recall quantifies the proportion of actual positives that are correctly identified, reflecting the coverage of positive cases — both are particularly significant in medical applications.

For the overall performance assessment, we adopted the F1 score, which is the harmonic mean of Precision and Recall. Additionally, AUC-ROC was used to evaluate the model's ability to distinguish between positive and negative samples. The ROC curve plots the True Positive Rate (vertical axis) against the False Positive Rate (horizontal axis), providing a classification-independent measure of performance.

**Table 2. Experimental results of different models**

| Model | Mean Acc | Precision | Recall | F1 Score | AUC |
|---|---|---|---|---|---|
| GBRT | 0.72 | 0.74 | 0.67 | 0.68 | 0.80 |
| KNN | 0.78 | 0.86 | 0.58 | 0.66 | 0.84 |
| DNN | 0.83 | 0.88 | 0.77 | 0.79 | 0.87 |
| DT | 0.68 | 0.69 | 0.67 | 0.66 | 0.68 |
| SVM | 0.80 | 0.87 | 0.70 | 0.74 | 0.88 |
| XGBoost | 0.67 | 0.66 | 0.58 | 0.60 | 0.71 |
| **BDPM** | **0.97** | **0.97** | **0.95** | **0.96** | **0.97** |

A higher AUC value indicates better predictive capability, with values ≥ 0.85 generally considered acceptable for medical models. The calculation formulas are as follows:

$$\text{Accuracy} = \frac{TP + TN}{TP + TN + FP + FN} \quad (20)$$

$$\text{Precision} = \frac{TP}{TP + FP} \quad (21)$$

$$\text{Recall} = \frac{TP}{TP + FN} \quad (22)$$

$$F1 = 2 \times \frac{\text{Precision} \times \text{Recall}}{\text{Precision} + \text{Recall}} \quad (23)$$

$$TPR = \frac{TP}{TP + FN} \quad (24)$$

$$FPR = \frac{FP}{FP + TN} \quad (25)$$

$$AUC = \int_0^1 TPR(FPR)\,d(FPR) \quad (26)$$

The definitions of the evaluation metrics are as follows:

TP (True Positives): Number of healthy samples correctly predicted by the model.

TN (True Negatives): Number of diseased samples correctly predicted by the model.

FP (False Positives): Number of diseased samples incorrectly predicted as healthy.

FN (False Negatives): Number of healthy samples incorrectly predicted as diseased.

### 3.3.1. Results of comparative experiments

To systematically evaluate the performance of the BDPM method in comparison with other commonly used classification algorithms for predicting PD based on gut microbiota data, this study trained and tested all models using the same dataset and parameter settings. We summarized the results in Table 2.

As shown in Table 2, the BDPM method outperforms the other classification models across multiple evaluation metrics. To further illustrate the differences among the five performance indicators, we present a corresponding heatmap in Fig. 4. In addition, the ROC curves for all methods are displayed in Fig. 5.

As shown in the results, the proposed BDPM method outperforms other algorithms in both accuracy and reliability. Specifically, it achieves an accuracy of 0.97, precision of 0.97, recall of 0.95, F1 score of 0.96, and AUC of 0.97. This systematic comparison confirms the superior performance of BDPM in processing gut microbiota data and highlights its significant advantages over existing classification methods.

### 3.3.2. Ablation Study

This study compares BDPM with representative classification algorithms. To comprehensively evaluate the contribution of each module in BDPM, we also designed ablation experiments.

We adopted a controlled experimental framework in which key components were sequentially removed from the baseline model to quantify their impact. Specifically, we examined whether data normalization, feature importance scoring using Random Forest (RF), Recursive Feature Elimination (RFE), embedding of a self-attention mechanism, and integration with SVM to leverage machine learning advantages — all influenced the performance of BDPM.

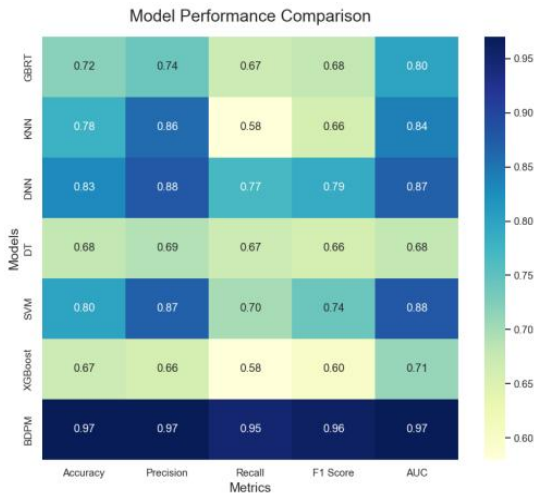

**Fig. (4).** The heatmap in Figure 4 visually confirms BDPM's superior performance over six benchmark algorithms, with marked improvements across all evaluated criteria including Mean Accuracy, Precision, Recall, F1 Score and AUC.

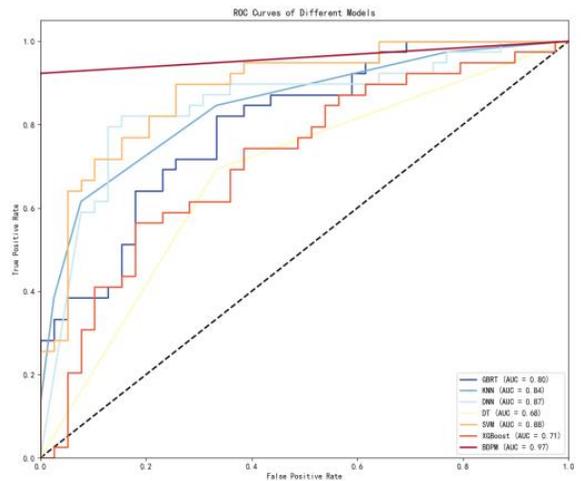

**Fig. (5).** The aggregated ROC plot (Fig. 5) shows BDPM consistently dominating other algorithms across all thresholds, with its curve occupying the upper-left quadrant most prominently.

**Table 3. Ablation Experiment Configurations**

| No. | Experimental Description |
|---|---|
| 1 | Omit normalization step |
| 2 | Remove SVM module |
| 3 | Remove attention mechanism |
| 4 | Remove RFE module |
| 5 | Replace RF with logistic regression |
| 6 | Baseline model (BDPM) |

**Table 4. Results of the Ablation Study**

| No. | Mean Acc | Precision | Recall | F1 Score |
|---|---|---|---|---|
| 1 | 0.91 | 0.88 | 0.95 | 0.91 |
| 2 | 0.96 | 0.96 | 0.92 | 0.94 |
| 3 | 0.95 | 0.95 | 0.95 | 0.95 |
| 4 | 0.95 | 0.95 | 0.95 | 0.95 |
| 5 | 0.93 | 0.90 | 0.95 | 0.92 |
| 6 | **0.97** | **0.97** | **0.95** | **0.96** |

**Table 5. Experimental Results of Bio-threshold Filtering**

| Bio-threshold | Mean Acc | Precision | Recall | F1 Score |
|---|---|---|---|---|
| 0.100% | 0.85 | 0.95 | 0.76 | 0.84 |
| 0.050% | 0.90 | 0.92 | 0.87 | 0.90 |
| 0.010% | 0.95 | 0.93 | 0.92 | 0.92 |
| 0.005% | **0.97** | **0.97** | **0.95** | **0.96** |
| 0.001% | 0.93 | 0.96 | 0.92 | 0.94 |

**Table 6. Experimental Results of Feature Number Selection**

| Num Features | Mean Acc | Precision | Recall | F1 Score |
|---|---|---|---|---|
| 20 | 0.85 | 0.83 | 0.87 | 0.85 |
| 25 | 0.92 | 0.90 | 0.89 | 0.90 |
| 30 | 0.94 | 0.94 | 0.93 | 0.93 |
| 35 | 0.96 | 0.95 | 0.93 | 0.94 |
| 40 | **0.97** | **0.97** | **0.95** | **0.96** |

The full BDPM model served as the baseline, and core components were either removed or modified in each ablation setting. The detailed experimental design is presented in Table 3.

Experiment 1 ignores the normalization processing step and directly uses the original strain abundance values; in Experiment 2, the results output from the LSTM output layer are directly used without using the decision boundary optimization of SVM; in Experiment 3, the attention mechanism embedded in the LSTM is removed, and only the infrastructure is retained; in Experiment 4, the results of the Random Forest are directly used; and in Experiment 5, logistic regression is used instead of the Random Forest.

The same dataset is used for validation with consistent hyperparameters, and the experimental results are shown in Table 4.

As shown in Table 4, the structural design of BDPM is well-justified, and the proposed method benefits from each module contributing positively to the overall model performance.

A comparison between Experiment 1 and Experiment 6 indicates that addressing scale differences in gut microbiota data is crucial; otherwise, it may significantly affect model performance. When comparing Experiment 2 with Experiment 6, it is evident that the dual-classifier LAS framework — based on a cascaded LSTM-SVM architecture constructed in the temporal-spatial phase — significantly enhances model performance compared to using a single classifier. The comparison between Experiment 3 and Experiment 6 confirms that the self-attention mechanism helps the LSTM classifier better capture dependencies among microbial species. Finally, comparing Experiments 4 and 5 with Experiment 6 shows slight improvements across all metrics, suggesting that feature processing not only simplifies the model but also boosts its performance.

### 3.3.3. Experiment on Bio-threshold

High-dimensional features in gut microbiota data can lead to the "curse of dimensionality" and introduce noise, which may negatively affect model performance. Moreover, significant abundance differences exist among microbial species. When these differences are substantial, features with very low relative abundances can be considered biologically insignificant and may be safely ignored to improve training accuracy.

Therefore, in the bio-deep dimensionality reduction stage, we applied a biological threshold filtering strategy based on the "rare species" theory in microbiomics. To evaluate the impact of different thresholds on model performance, we conducted five experiments using cutoffs of 0.1%, 0.05%, 0.01%, 0.005%, and 0.001%. The results are summarized in Table 5.

Our experiments on biological threshold selection validated the influence of threshold settings and demonstrated the rationality of the bio-deep dimensionality reduction approach. In Fig. 6, the optimal prediction performance was achieved when the threshold was set to 0.005%.

When the threshold was lower than 0.005%, accuracy and other metrics improved slightly, but precision decreased. When the threshold exceeded 0.005%, all performance metrics declined. Therefore, the selection of an appropriate biological threshold should be based on both the original data and ecological knowledge.

### 3.3.4. Experiment on the Impact of Feature Quantity

The number of selected features significantly impacts model performance in machine learning. Too many features may lead the model to capture noise, while too few can result in the loss of important information, thereby negatively affecting key performance metrics such as accuracy. Therefore, selecting an appropriate number of features helps improve training efficiency while maintaining model performance.

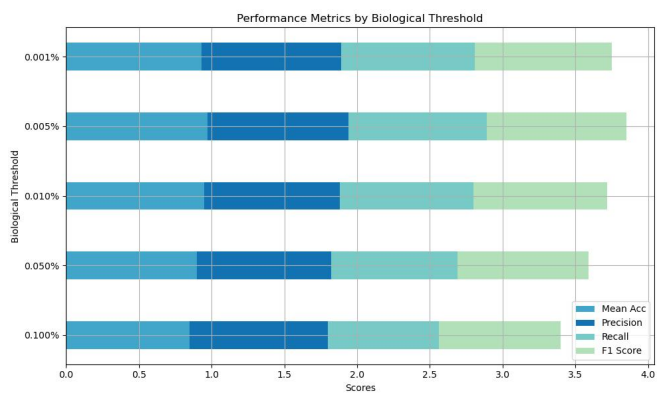

**Fig. (6).** In the biological threshold experiment, BDPM demonstrated optimal performance at the 0.005% threshold, while concurrently validating the feasibility of this interdisciplinary approach.

In this study, we investigated how feature affects model performance by varying the number of input features. Specifically, we tested six different feature counts: 20, 25, 30, 35, 40, and 45. The corresponding model performances were evaluated and are summarized in Table 6.

As shown in Fig. 7, the model achieves an accuracy of 0.85 with 20 features, indicating acceptable performance with room for improvement. When the number of features increases to 40, the model reaches its peak performance with an accuracy of 0.97. This suggests that, within a certain range, increasing the number of features enables the model to better capture key data associations and improve training effectiveness.

However, beyond the optimal feature count, performance begins to decline. At 45 features, the accuracy drops to 0.94. While increasing the number of features can enhance the model's learning capability, excessive features may introduce redundancy and ultimately degrade performance.

## 4. DISCUSSION

We successfully developed a predictive model for Parkinson's disease based on differential gut microbiota. By analyzing intestinal data from both healthy individuals and patients, the model demonstrated strong performance in terms of sensitivity and accuracy, further supporting the brain-gut axis theory.

During the study, we integrated biological theories to validate the impact of biological thresholds on Parkinson's disease, reflecting an interdisciplinary approach. By considering multiple factors, this work provides new insights for future research in this field.

Nevertheless, several limitations remain. First, the model relies on publicly available datasets with relatively small sample sizes, making it difficult to establish independent training and validation sets. As a result, its stability and generalizability require further improvement. Second, the dataset does not account for variables such as gender and nationality, which may influence outcomes and should be considered in future studies. Additionally, the abundance of gut microbiota varies significantly across stages, and their temporal dynamics warrant deeper investigation.

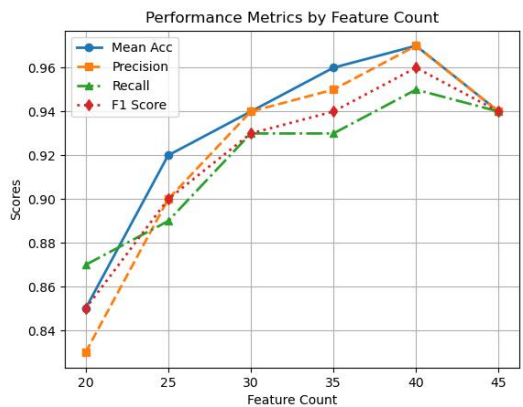

**Fig. (7).** It presents line charts comparing the experimental results for five different feature-number groups, revealing the existence of an optimal feature number.

In summary, this study presents a novel network model capable of predicting Parkinson's disease using differential gut microbiota, reinforcing the link between gut flora and the disease. Compared with previous approaches that employed single classifiers such as random forest[21, 23] and artificial neural networks[24], our dual-classifier framework demonstrates improved accuracy. Moreover, in contrast to earlier dual-classifier models proposed by Yu Bo et al.[25], our method incorporates the concept of biological thresholding, identifies optimal feature numbers, and proposes a more innovative and microbiome-applicable feature preprocessing and classification architecture. These findings offer a promising direction for the early detection and prevention of Parkinson's disease.

## CONCLUSION

Current approaches to gut microbiota analysis often overlook complex microbial interactions and may reduce predictive accuracy. Moreover, multi-level feature reduction strategies remain underexplored, despite the robustness of ML methods. To address these issues, we propose BDPM, a classifier network based on the gut-brain axis theory for PD. BDPM improves performance through an enhanced architecture and novel feature processing. We build a full pipeline from preprocessing to classification, offering a reference framework for similar tasks.

Despite these advances, our method has limitations in terms of data scale and diversity. In future work, we will explore temporal dynamics among microbial species and expand clinical data collection to improve model reliability and clinical applicability. Furthermore, we aim to refine the algorithm and extend its application to other neurodegenerative diseases influenced by gut microbiota, such as Alzheimer's disease, thereby promoting interdisciplinary research at the intersection of microbiology and neurology.

## ETHICS APPROVAL AND CONSENT TO PARTICIPATE

Not applicable

## HUMAN AND ANIMAL RIGHTS

Not applicable

## RESEARCH INVOLVING HUMANS

Not applicable

## AVAILABILITY OF DATA AND MATERIALS

This study utilized a publicly available dataset, with gut microbiome data sourced from L. Mao, accessible at: https://www.ncbi.nlm.nih.gov/

## FUNDING

No.

## CONFLICT OF INTEREST

The authors declare no conflict of interest, financial or otherwise.

## ACKNOWLEDGEMENTS

Declared none.

## REFERENCES

[1] Y. Ben-Shlomo, S. Darweesh, J. Llibre-Guerra, et al., The epidemiology of Parkinson's disease, The Lancet 403(10423) (2024) 283-292.

[2] E.R. Dorsey, R. Constantinescu, J.P. Thompson, et al., Projected number of people with Parkinson disease in the most populous nations, 2005 through 2030, Neurology 68(5) (2007) 384-6.

[3] E.R. Dorsey, T. Sherer, M.S. Okun, et al., The Emerging Evidence of the Parkinson Pandemic, Journal of Parkinson's Disease 8(s1) (2018) S3-S8.

[4] H. Nishiwaki, M. Ito, T. Hamaguchi, et al., Short chain fatty acids-producing and mucin-degrading intestinal bacteria predict the progression of early Parkinson's disease, npj Parkinson's Disease 8(1) (2022).

[5] G. Rizzo, M. Copetti, S. Arcuti, et al., Accuracy of clinical diagnosis of Parkinson disease, Neurology 86(6) (2016) 566-576.

[6] T. Simuni, A. Siderowf, S. Lasch, et al., Longitudinal Change of Clinical and Biological Measures in Early Parkinson's Disease: Parkinson's Progression Markers Initiative Cohort, Movement Disorders 33(5) (2018) 771-782.

[7] D. Berg, P. Borghammer, S.-M. Fereshtehnejad, et al., Prodromal Parkinson disease subtypes — key to understanding heterogeneity, Nature Reviews Neurology 17(6) (2021) 349-361.

[8] S.-M. Fereshtehnejad, C. Yao, A. Pelletier, et al., Evolution of prodromal Parkinson's disease and dementia with Lewy bodies: a prospective study, Brain 142(7) (2019) 2051-2067.

[9] [9] H.R. Morris, M.G. Spillantini, C.M. Sue, et al., The pathogenesis of Parkinson's disease, The Lancet 403(10423) (2024) 293-304.

[10] B. Huang, S.W.H. Chau, Y. Liu, et al., Gut microbiome dysbiosis across early Parkinson's disease, REM sleep behavior disorder and their first-degree relatives, Nature Communications 14(1) (2023).

[11] S. Kim, S.-H. Kwon, T.-I. Kam, et al., Transneuronal Propagation of Pathologic α-Synuclein from the Gut to the Brain Models Parkinson's Disease, Neuron 103(4) (2019) 627-641.e7.

[12] Y. Qian, X. Yang, S. Xu, et al., Gut metagenomics-derived genes as potential biomarkers of Parkinson's disease, Brain 143(8) (2020) 2474-2489.

[13] S. Romano, J. Wirbel, R. Ansorge, et al., Machine learning-based meta-analysis reveals gut microbiome alterations associated with Parkinson's disease, Nature Communications 16(1) (2025).

[14] I. El Maachi, G.-A. Bilodeau, W. Bouachir, Deep 1D-Convnet for accurate Parkinson disease detection and severity prediction from gait, Expert Systems with Applications 143 (2020).

[15] S. Sivaranjini, C.M. Sujatha, Deep learning based diagnosis of Parkinson's disease using convolutional neural network, Multimedia Tools and Applications 79(21-22) (2019) 15467-15479.

[16] J. Mei, C. Desrosiers, J. Frasnelli, Machine Learning for the Diagnosis of Parkinson's Disease: A Review of Literature, Frontiers in Aging Neuroscience 13 (2021).

[17] D. Palacios-Alonso, G. Melendez-Morales, A. Lopez-Arribas, et al., MonParLoc: A Speech-Based System for Parkinson's Disease Analysis and Monitoring, IEEE Access 8 (2020) 188243-188255.

[18] S. Gao, Z. Wang, Y. Huang, et al., Early detection of Parkinson's disease through multiplex blood and urine biomarkers prior to clinical diagnosis, npj Parkinson's Disease 11(1) (2025).

[19] K. Tsukita, H. Sakamaki-Tsukita, S. Kaiser, et al., High-Throughput CSF Proteomics and Machine Learning to Identify Proteomic Signatures for Parkinson Disease Development and Progression, Neurology 101(14) (2023).

[20] E. Pantaleo, A. Monaco, N. Amoroso, et al., A Machine Learning Approach to Parkinson's Disease Blood Transcriptomics, Genes 13(5) (2022).

[21] L. Mao, Y. Zhang, J. Tian, et al., Cross-Sectional Study on the Gut Microbiome of Parkinson's Disease Patients in Central China, Frontiers in Microbiology 12 (2021).

[22] D. Berg, R.B. Postuma, C.H. Adler, et al., MDS research criteria for prodromal Parkinson's disease, Movement Disorders 30(12) (2015) 1600-1611.

[23] F. Clasen, S. Yildirim, M. Arıkan, et al., Microbiome signatures of virulence in the oral-gut-brain axis influence Parkinson's disease and cognitive decline pathophysiology, Gut Microbes 17(1) (2025).

[24] M. Boodaghidizaji, T. Jungles, T. Chen, et al., Machine learning based gut microbiota pattern and response to fiber as a diagnostic tool for chronic inflammatory diseases, BMC Microbiology 25(1) (2025).

[25] B. Yu, H. Zhang, M. Zhang, Deep learning-based differential gut flora for prediction of Parkinson's, Plos One 20(1) (2025).